\documentclass[lettersize,journal]{IEEEtran}
\usepackage{amsmath,amsfonts}
\usepackage{algorithmic}
\usepackage{algorithm}
\usepackage{array}
\usepackage{subcaption}
\usepackage{xcolor}
\usepackage{textcomp}
\usepackage{stfloats}
\usepackage{url}
\usepackage{verbatim}
\usepackage{graphicx}
\usepackage{cite}
\usepackage{amssymb}
\begin{document}

\title{Communication-Efficient Federated Learning via Regularized Sparse Random Networks}

\author{
\IEEEauthorblockN{Mohamad Mestoukirdi, Omid Esrafilian, David Gesbert, Qianrui Li, Nicolas Gresset}

\thanks{\newline M. Mestoukirdi, N. Gresset are with Mitsubishi Electric R\&D Centre Europe. 
Email: \{M.Mestoukirdi,N.Gresset\}@fr.merce.mee.com
\newline M. Mestoukirdi, O. Esrafilian and D. Gesbert are with the Communication
Systems Department, EURECOM, Sophia-Antipolis, France. 
Emails: \{mestouki, esrafili, gesbert\}@eurecom.fr.
\newline Qianrui Li was with Mitsubishi Electric R\&D Centre Europe during the
development of this work, and is now with CICT Mobile Communications Technology Co. Ltd., Beijing 100832 and with the State Key Laboratory of Wireless Mobile Communications, China Academy of Telecommunications Technology (CATT), Beijing 100191. 
Email: qianrui\_li@qq.com
\newline Part of the work by Omid Esrafilian was funded via the HUAWEI
France supported Chair on Future Wireless Networks at EURECOM.
}}
\markboth{Journal of \LaTeX\ Class Files,~Vol.~14, No.~8, August~2021}%
{Shell \MakeLowercase{\textit{et al.}}: A Sample Article Using IEEEtran.cls for IEEE Journals}

\IEEEpubid{~\copyright~ }

\maketitle

\begin{abstract}

This work presents a new method for enhancing communication efficiency in stochastic Federated Learning that trains over-parameterized random networks. In this setting, a binary mask is optimized instead of the model weights, which are kept fixed. The mask characterizes a sparse sub-network that is able to generalize as good as a smaller target network. Importantly, sparse binary masks are exchanged rather than the floating point weights in traditional federated learning, reducing communication cost to at most 1 bit per parameter (Bpp). We show that previous state of the art stochastic methods fail to find sparse networks that can reduce the communication and storage overhead using consistent loss objectives. To address this, we propose adding a regularization term to local objectives that acts as a proxy of the transmitted masks entropy, therefore encouraging sparser solutions by eliminating redundant features across sub-networks. Extensive empirical experiments demonstrate significant improvements in communication and memory efficiency of up to five magnitudes compared to the literature, with minimal performance degradation in validation accuracy in some instances.
\end{abstract}


\section{Introduction}
\IEEEPARstart{F}{ederated} Learning (FL) was introduced to address privacy concerns associated with centrally training large models that rely on raw data from edge devices \cite{ref1}. It enables clients to collaboratively train models in a distributed way under the supervision of a parameter server (PS). This approach involves iterative training executed over several communication rounds of decentralized on-device model optimization and
centralized aggregation of model updates. While FL ensures data privacy, it faces challenges in terms of communication efficiency due to the large size of exchanged model updates during each communication round for both uplink (UL) and downlink (DL). Recent efforts have focused on reducing communication overhead leveraging compression via quantization and model sparsification techniques \cite{ref2,ref3} on the exchanged model weights. Despite these efforts, exchanged compressed models are still represented according to float bit-representations (e.g., 32/16 bits per model weight), leading to significant communication overhead as the size of trained models increases (e.g., LLM).

A recent work \cite{ref4} has revealed that in over-parameterized random neural networks, it is possible to find smaller sub-networks that perform just as well as a fully trained target network in terms of generalization. These sub-networks are produced by element-wise multiplication of a sparse binary mask with the initial weights of the over-parameterized network (i.e while fixing the weights). In this case, the binary mask is optimized to identify the initial weights that would constitute a sub-network with similar generalization performance as the target network. Subsequently, the authors in \cite{ref5} leverage the subset-sum approximation problem \cite{ref6} to prove the existence of those sub-networks. They show that dense target networks with width $d$ (neuron count in a layer) and depth $l$ (number of layers) can be closely approximated by pruning an over-parameterized dense random network with a width $O(\log(dl))$ times larger and a depth twice as deep. This discovery is particularly interesting for FL training, due to the lower communication overhead associated with exchanging binary masks in the UL and DL instead of float-bit representations of the weight updates. In \cite{ref7}, the authors introduce FedMask, a personalized Federated Learning (FL) algorithm based on pruning over-parameterized random networks.  FedMask is a deterministic algorithm that involves pruning a random network by optimizing personalized binary masks using Stochastic Gradient Descent (SGD), aiming to approximate the personalized target networks that fit the heterogeneous datasets found at the devices. Their approach has been shown to ensure a 1-bit-per-parameter (1Bpp) communication cost per each round of communication to exchange the updates in FL training. This can be attributed to the nature of their algorithm, which optimizes the binary masks within a constrained search space, wherein the masks demonstrate an equiprobable occurrence of ones and zeros. Recently, a stochastic approach called FedPM \cite{ref8} was introduced as an alternative to the deterministic FedMask. FedPM requires edge devices to identify a global probability mask, as apposed to the deterministic mask in FedMask. Binary masks then are sampled from the global probability mask, characterizing sub-networks with strong generalization capabilities over the diverse datasets of the edge devices. The results of their approach demonstrate state-of-the-art accuracy and communication efficiency compared to FedMask and other baseline methods. However, we show that their method fails to discover sparse networks, leaving a significant amount of unnecessary redundancy in terms of the size of the found sub-networks. 

This work builds upon the foundation of stochastic masking techniques \cite{ref4,ref8}, leveraging their favorable generalization performance and convergence while aiming to enhance communication and memory efficiency. Our main contributions are summarized as follows:

\begin{itemize}

\item We introduce a new objective function leading to effectively restricting the search space to discover a limited set of sparse sub-networks within the over-parameterized random network. These sub-networks offer both communication efficiency and strong generalization performance in Federated learning settings compared to the literature.

\item Through simulations, we demonstrate that the non-structural sparsity enforced through the proposed loss function results in significantly sparser solutions compared to state-of-the-art algorithms such as FedPM. Importantly, this sparsity gain is achieved without sacrificing generalization, leading to up to about 5 magnitudes in terms of communication and memory efficiency during training. 

\end{itemize}

\section{System Model and Problem Formulation}
 We suppose that there are $K$ edge-devices in the system with datasets denoted by $\{\mathcal{D}_i\}^K_{i=1}$. The training procedure commences as the parameter server sends a randomly initialized network to the edge devices. This is accomplished by providing the devices with both the network's structure and an initialization seed, enabling them to construct the network's layers and weights. We denote the initialized weights of the network by $\boldsymbol{w}_{\text{init}} = \left(w_1,\cdots,w_n\right) \in \mathbb{R}^{n}$. The primary objective is to identify a global binary mask $\boldsymbol{{m}}$, yielding a sub-network $y_{{\boldsymbol{m}}}: \mathbb{R}^{n} \longrightarrow \mathbb{R}$ given according to\footnote{For ease of representation, we use a linear model in \eqref{eq1}.}: \begin{equation}
     y_{{\boldsymbol{m}}}(\boldsymbol{x}) = \left({\boldsymbol{m}}\otimes \boldsymbol{w}_{\text{init}}\right)^T\cdot \,\boldsymbol{x},
     \label{eq1}
 \end{equation}
where $\boldsymbol{x}\in \mathbb{R}^{n}$ denotes a feature vector, $\otimes$ denotes the element-wise multiplication operator, and $(\cdot)$ denotes vector multiplication. The produced sub-network minimizes the empirical risk function in accordance to:

\begin{equation}
    \min_{\boldsymbol{{m}}} F(\boldsymbol{{m}}) = \frac{1}{\sum_i |\mathcal{D}_i|}\sum^K_{k=1}|\mathcal{D}_k| \ell(y_{{\boldsymbol{m}}},\mathcal{D}_k),
    \label{objectivef}
\end{equation}
 
where $F(\boldsymbol{{m}})$ denotes the   empirical risk of $y_{\boldsymbol{{m}}}$ over the devices datasets. $\ell(.,.)$ denotes the local loss function and $|\mathcal{D}_k|$ the dataset size of device $k$.

We denote the target network that we aim at approximating by $y_{\text{target}}$. The number of sub-networks that can be found within the over-parameterized network to approximate $y_{\text{target}}$ increases with its size \cite{ref4,ref5,ref9}. Accordingly, constructing a sufficiently over-parameterized random network according to the rules derived in \cite{ref5} guarantees with a high probability that a sub-network $y$ exists, such that $y \approx y_{\text{target}}$. To this end, we aim at identifying the individual weights of $\boldsymbol{w}_{\text{init}}$ that play a role in producing sub-networks capable of generalizing as effectively as $y_{\text{target}}$. This is achieved by maximizing the likelihood of these weights while disregarding the weights that do not offer any meaningful contribution towards that objective. Akin to \cite{ref8}, along-side the initialized weights, the users receive a global probability mask vector\footnote{Initial global probability mask parameters can be randomly initialized (e.g. from $\mathcal{U}[0,1]).$} ${\boldsymbol{\theta}} \in [0,1]^n$:
\begin{equation}
    \boldsymbol{\theta}_i \longleftarrow \boldsymbol{\theta}(t)\\ 
\end{equation}
using the probability mask, each user $i$ derives a local score vector $\boldsymbol{s}_i$, according to : 
 \begin{equation}
      \boldsymbol{s}_i = \sigma^{-1}(\boldsymbol{\theta}_i)
 \end{equation}
where $\sigma(\boldsymbol{s})$ denotes the sigmoid function applied to each parameter of the vector $\boldsymbol{s }$. The probability mask represents the likelihood of each particular weight in $\boldsymbol{w}_{\text{init}}$ contributing to the chosen sub-network in \eqref{eq1}. Once each device $i$ receive the global probability mask $\boldsymbol{\theta(t)}$ from the server, training starts by sampling a binary mask which characterizes the local sub-network $y_{\boldsymbol{m}^h_i}$ to minimize its loss, given by
\begin{equation}
     y_{\boldsymbol{m}^h_i}(\boldsymbol{x}) = \left(\boldsymbol{m}^h_i\otimes \boldsymbol{w}_{\text{init}}\right)^T \cdot \,\boldsymbol{x}   \quad ,\,\boldsymbol{m}^h_i \sim \text{Bernoulli}(\boldsymbol{\theta}^h_i),
     \label{eq1}
 \end{equation}

Here $h$ denotes the local mini-batch iterations count, where $\boldsymbol{\theta}^{h=0}_i = 
\boldsymbol{\theta}(t)$. Similar to \cite{ref4}, Instead of directly optimizing $\boldsymbol{\theta}^h_i$, the score vector is employed in the optimization process. This ensures smooth and unbiased\footnote{For instance, FedMask relies on optimizing a deterministic mask via SGD, and then thresholding the resultant updated mask. The threshold operation results in biased updates.} updates of $\boldsymbol{\theta}$. The scores and probability masks are updated at $\emph{each}$ mini-batch iteration $h$ according to:

\begin{equation}
 \boldsymbol{{\theta}}^h_i = \sigma(\boldsymbol{s}^{h-1}_i - \frac{\eta}{|\mathcal{B}^h|} \nabla_{\boldsymbol{s}^{h-1}_i} \ell(y_{\boldsymbol{m}^{h-1}_i}, \mathcal{B}^h)),
\label{probability_update}
\end{equation}

where $\eta$ is the learning rate, $\mathcal{B}^h\subseteq \mathcal{D}_i$ is a mini-batch, and $|\mathcal{B}^h|$ denotes its' cardinality. $\nabla_{\boldsymbol{s}^{h-1}_i} \ell(y,\mathcal{B}^h)$ denotes the gradient of the loss function (i.e the cross entropy loss in classification tasks) of the local sub-network $y_{\boldsymbol{m}^{h-1}_i}$ -- sampled during the current iteration $h$ -- over the mini-batch $\mathcal{B}^h$ at device $i$, with respect to the scores vector $\boldsymbol{s}^{h-1}_i$. Accordingly, each parameter indexed $j$ of the score vector $\boldsymbol{s}^{h-1}_{i,j}$ are optimized locally using the chain rule according to:
\begin{equation}
    \boldsymbol{s}^h_{i,j} = \boldsymbol{s}^{h-1}_{i,j} - \eta\left(\frac{\partial \ell}{\partial y_{\boldsymbol{m}^{h-1}_i}} \times \frac{\partial y_{\boldsymbol{m}^{h-1}_i}}{\partial {m}^{h-1}_{i,j}}\times \frac{\partial {m}^{h-1}_{i,j}}{\partial {\theta}^{h-1}_{i,j}}\times \frac{\partial {\theta}^{h-1}_{i,j}}{\partial {s}^{h-1}_{i,j}}\right).
\end{equation}

$m^{h-1}_{i,j}$ and $ {\theta}^{h-1}_{i,j}$ denote the $j^{\text{th}}$ parameters of $\boldsymbol{m}^{h-1}_i$ and $\boldsymbol{\theta}^{h-1}_i$ respectively. We omit the local iteration count $h$ in the following expressions for ease of representation. Note that the sampling operation $m^h_{i,j} \sim \text{Bernoulli}(\theta^h_{i,j})$ is not differentiable. Therefore $\frac{\partial {m}^h_{i,j}}{\partial {\theta}^h_{i,j}}$ can be approximated using straight-through estimators \cite{ref4,ref8}. 
Next, after optimizing the scores for a  number of local iterations, let $\boldsymbol{\hat{\theta}}_i(t)$ denote the locally produced probability mask at round $t$. For each client $i$, a binary mask $\boldsymbol{\hat{m}}_i$ is sampled according to:
\begin{equation*}
\hat{\boldsymbol{m}}_i(t) \sim \text{Bernoulli}(\hat{\boldsymbol{\theta}_i}(t)).
\end{equation*}

 These binary masks are then sent to the server. The sent masks highlight the weights contributing to the best sub-networks. This approach effectively reduces the communication cost (entropy) to a maximum of 1 bit per parameter (1Bpp), where the actual entropy depends on the sparsity of the mask. The server then performs averaging to generate a global probability mask according to:

\begin{equation}
\boldsymbol{\theta}(t+1) \longleftarrow \frac{1}{\sum_k|\mathcal{D}_k|}\sum_i |\mathcal{D}_i| \hat{\boldsymbol{m}}_i(t).
\label{global_mask}
\end{equation}

The resultant global probability mask $\boldsymbol{\theta}(t+1)$ is re-distributed to the devices in the DL to commence the next communication round. The global mask has been demonstrated in \cite{ref8} to be an unbiased estimate of the true global probability mask $\boldsymbol{\bar{\theta}}$, which is given by $\boldsymbol{\bar{\theta}}(t+1) = \frac{1}{\sum_k|\mathcal{D}_k|}\sum_i |\mathcal{D}_i|  \hat{\boldsymbol{\theta}}_i(t)$.

\section{Intuition and Proposed Loss}
\subsection{Intuition}
 We  delineate the shortcomings of the current state-of-the-art technique in FedPM \cite{ref8} with regards to the sparsity level of the networks identified, by conducting a comprehensive evaluation of the original optimization algorithm employed. This evaluation is conducted from the vantage point of each individual learner, under the premise of an absence of regularization in the loss function.

During FedPM training, within every local iteration (e.g. mini-batch update), individual devices sample a distinct instance sub-network based on a received probability mask as outlined in \eqref{eq1}. As a result of the considerable scale of the over-parameterized random network, the sampled sub-networks may be entirely new for the devices at each local iteration. Subsequently, each device calculates the loss specific to the sampled network and then back-propagates the gradients to minimize the loss. This is done by adjusting the scores in directions that activate or deactivate the fixed random weights appropriately. In subsequent local iterations, additional networks are sampled, and their weight scores are tuned to minimize their corresponding loss. From a broader perspective, the local stochastic sub-network sampling step designed in FedPM implicitly promotes the minimization of the  average loss of  sub-networks sampled from the probability mask at each device, given by:
\begin{equation}
\label{redundants}
    \frac{1}{H} \sum^H_{h=1} \ell(y_{{\boldsymbol{m}^h_i}},\mathcal{B}_h), \quad
    \boldsymbol{m}^h_i \sim \text{Bernoulli}(\boldsymbol{\theta}^h_i)
\end{equation}where $H$ denotes the number of local mini-batch updates during each round. Due to the substantial number of existing sub-networks that can generalize well, this sampling step results in redundancy in terms of the number of optimized sub-networks and accordingly the number of activated weights, which reduces the sparsity of the transmitted masks.  Additionally, due to the asymptotic behavior of the Sigmoid function near saturation, pruning the redundant sub-networks optimized locally according to \eqref{redundants} proves challenging. Specifically, as the parameters scores approach asymptotic maximum values, the derivatives (given by $\frac{d\theta}{ds} = \sigma(s)(1-\sigma(s))$) converge to zero. Subsequently, the gradients of the loss function with respect to redundant subnetwork parameters diminish. Herein, the incorporation of an additional regularization term helps counteract the gradient vanishing, enabling redundant parameters to be pruned over time despite the challenges of the sigmoid's flat extremes.

% Note that given the flatness of the sigmoid function on the extremes, the redundant sub-networks chosen by the different devices, proves hard to eliminate. This stems from the fact that the derivative, given by $\frac{d\theta}{\theta}$, converges to zero, as the score increase indicating the importance of choosing a certain parameter, and ultimately the gradient of the loss with respect to the redundant sub-network parameters converges towards zero. That said, the regularization term offers this counter action towards reviving the gradients of the scores of the parameters of redundant networks to allow them to be phased out, yielding a sparse final sub-network. 
\subsection {Proposed Objective and Loss functions}
To mitigate the aforementioned limitations, we propose a new objective incorporating a regularization term that acts as a proxy for the entropy of the binary vectors transmitted in the uplink by each device, together with the original loss between the predicted output and the ground truth label. The new objective which we aim at solving can be given as follows: 
\begin{align}
&\min_{\boldsymbol{\theta}} \bar{F}(\boldsymbol{{m}}) = \frac{1}{\sum_i |\mathcal{D}_i|}\sum^K_{k=1}|\mathcal{D}_k| \ell(y_{{\boldsymbol{m}}},\mathcal{D}_k) + \frac{\lambda}{n} H(\boldsymbol{m}),
    \label{newobj}
\end{align}
 where $\boldsymbol{m} \sim \text{Bern}(\boldsymbol{\theta})$, $\boldsymbol{\theta}$ is the global probability mask given in  \eqref{global_mask}, $\lambda$ is a regularization term and { $H(\boldsymbol{m})$ is the entropy of the global mask, given by :
\begin{equation}
H(\boldsymbol{m}) = \frac{1}{K}\sum_k H(\boldsymbol{m}_k) = -\frac{1}{K}\sum_k p_{k,0}\log(p_{k,0}) + p_{k,1}\log(p_{k,1}),
\end{equation}
where $\boldsymbol{m}_k$ is the mask transmitted in the UL by device $k$. Herein, local mask binary elements are seen to be generated by a device specific binary source with probabilities $p_{0,k}$ and $p_{1,k}$ respectively, for each device $k$.  }
\begin{figure*}[ht!]
\centering

    \begin{subfigure}[t]{0.31\textwidth}
         \centering
         \includegraphics[width=\textwidth]{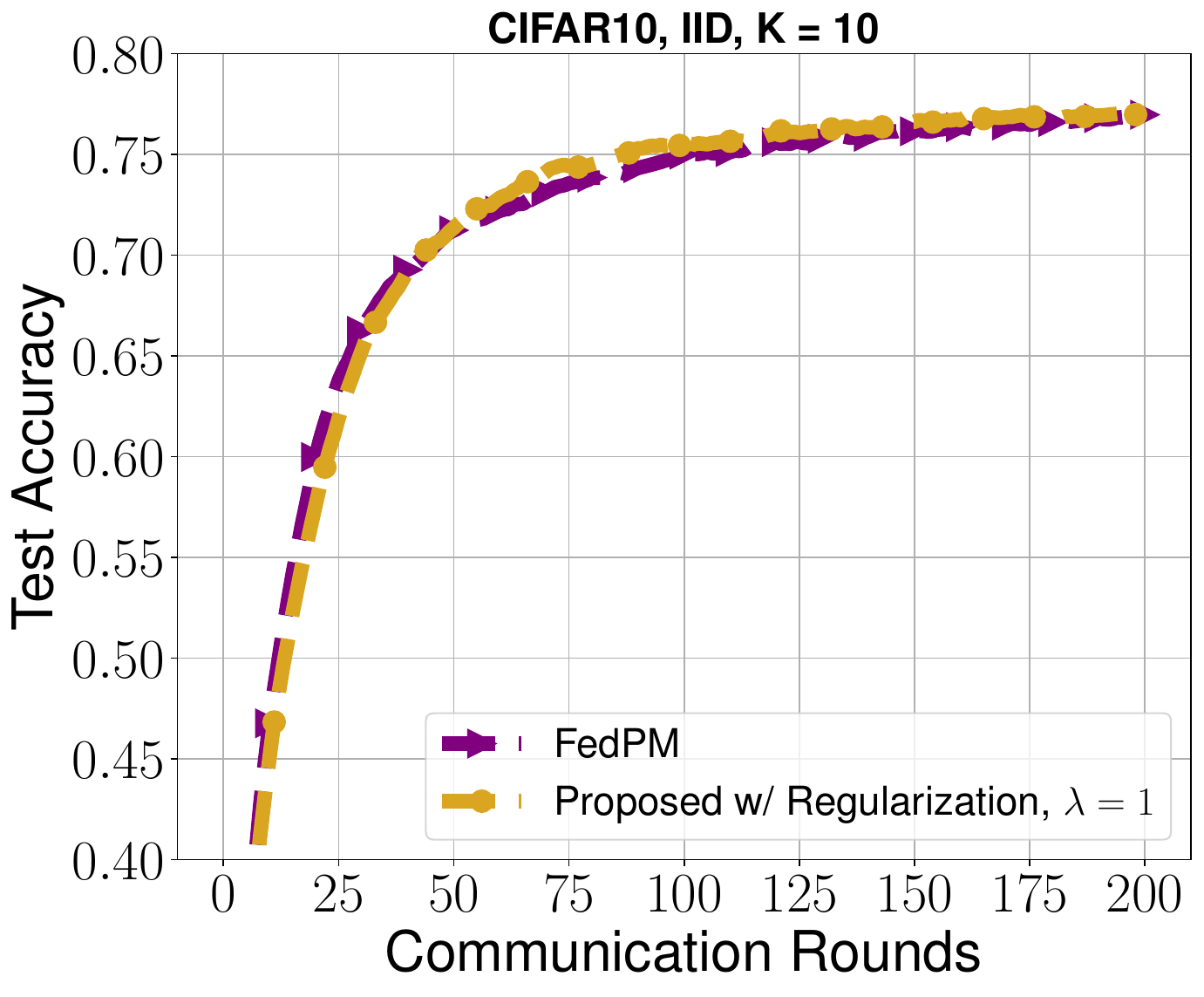}
         \centering
         \captionsetup{labelformat=empty}
         \caption{}

     \end{subfigure}
      \begin{subfigure}[t]{0.31\textwidth}
         \centering
    \includegraphics[width=\textwidth]{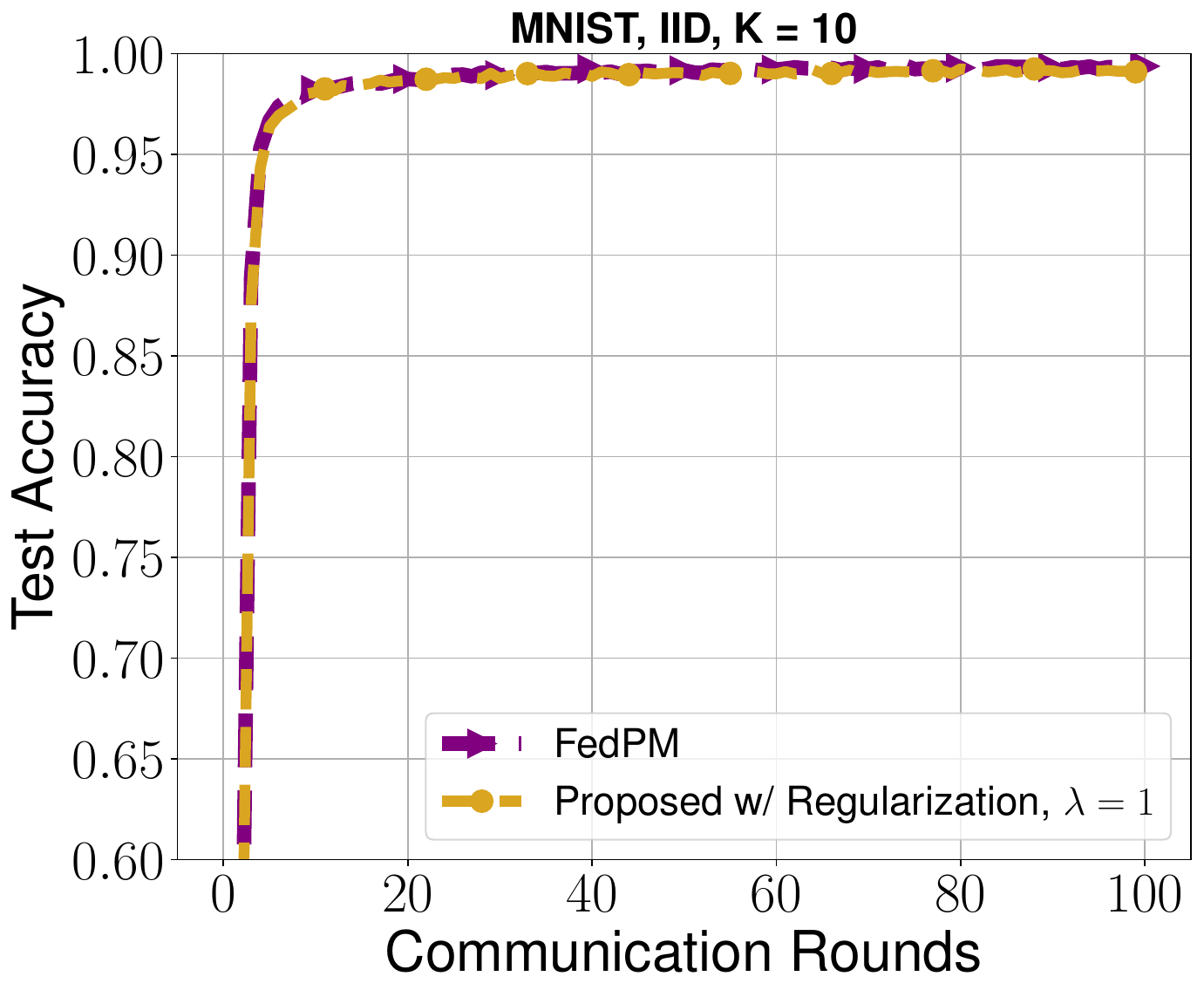}
    \captionsetup{labelformat=empty}
         \caption{}
         \label{fig:EMNIST_cov}
     \end{subfigure}
           \begin{subfigure}[t]{0.31\textwidth}
         \centering
    \includegraphics[width=\textwidth]{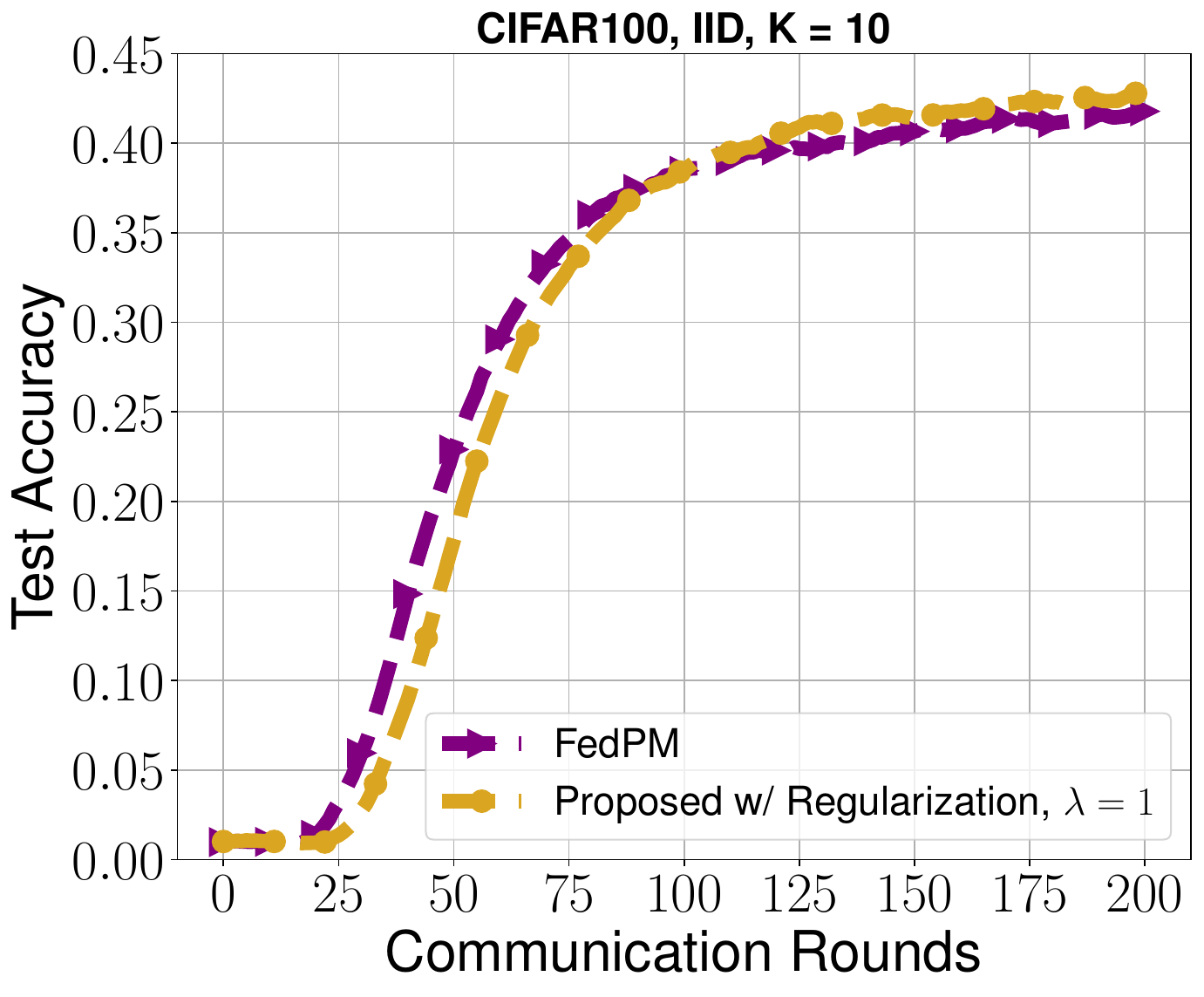}
        \captionsetup{labelformat=empty}
         \caption{}
         \label{fig:EMNIST_cov}
     \end{subfigure}
     
    \begin{subfigure}[t]{0.31\textwidth}
         \centering
         \includegraphics[width=\textwidth]{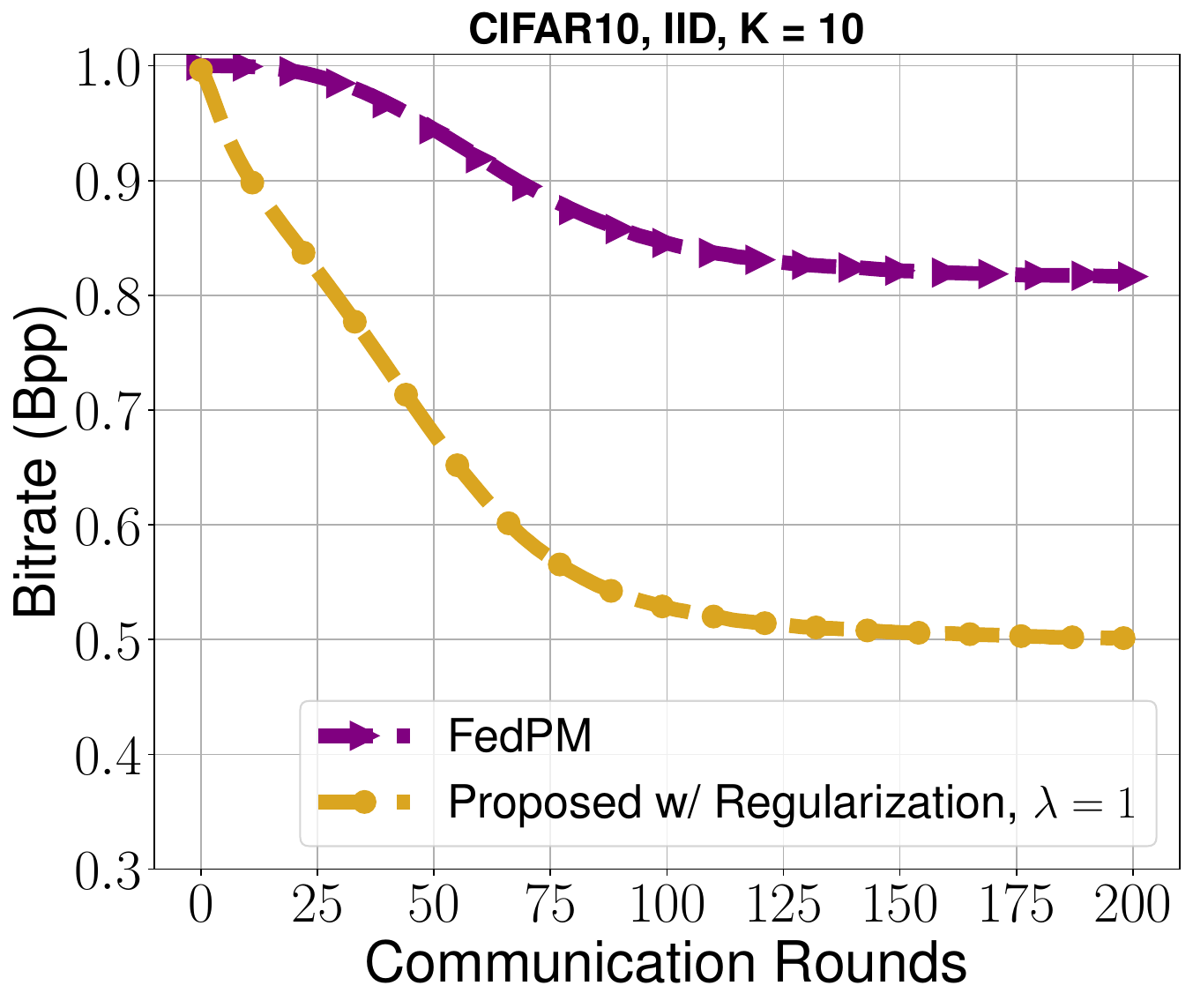}
         \centering
         \captionsetup{labelformat=empty}
         \caption{(a) CIFAR10, IID}
         \label{fig:EMNIST_label}
     \end{subfigure}
      \begin{subfigure}[t]{0.31\textwidth}
         \centering
    \includegraphics[width=\textwidth]{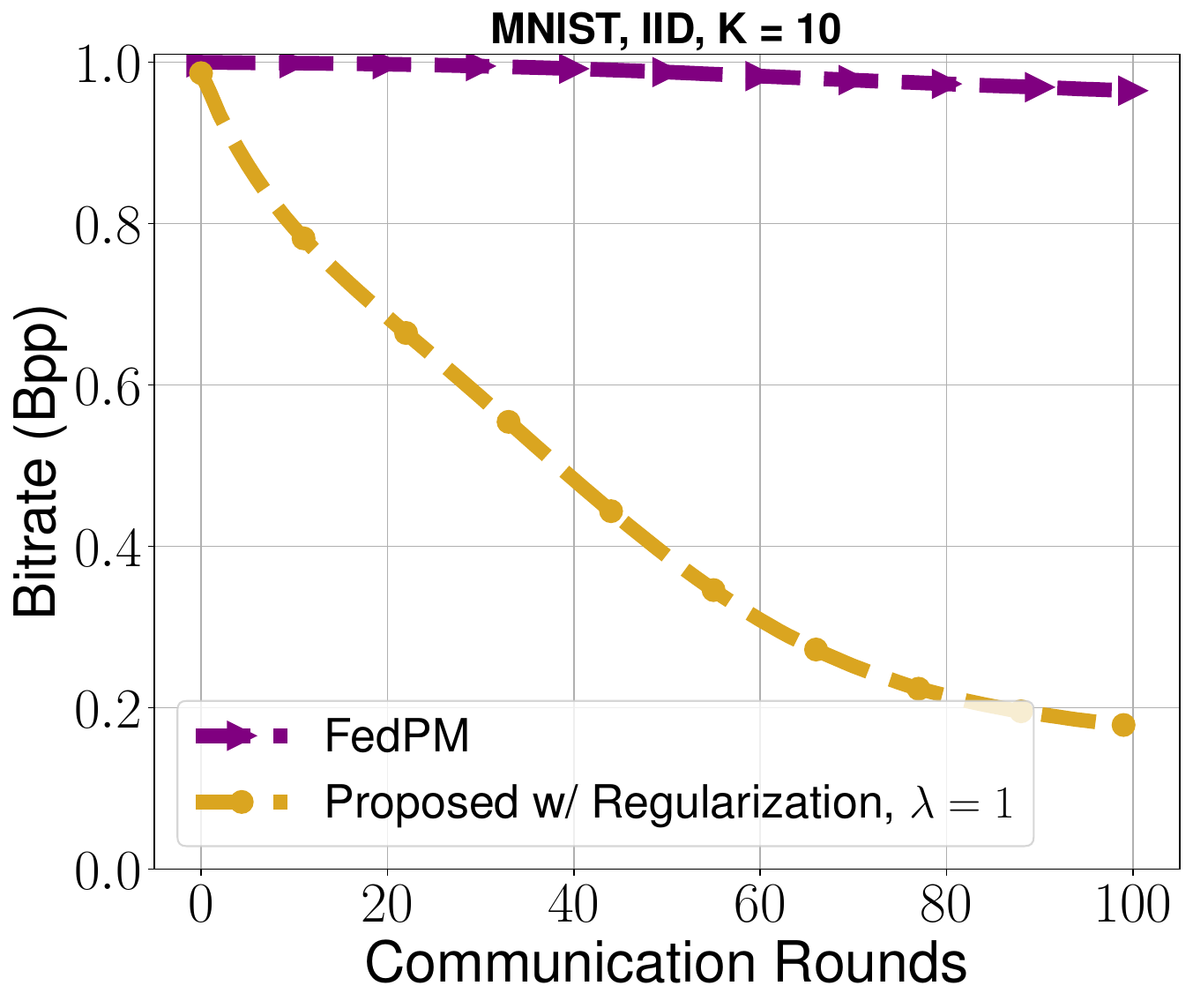}
    \captionsetup{labelformat=empty}
    \caption{(b) MNIST, IID}
         \label{fig:EMNIST_cov}
     \end{subfigure}
\begin{subfigure}[t]{0.31\textwidth}
         \centering
    \includegraphics[width=\textwidth]{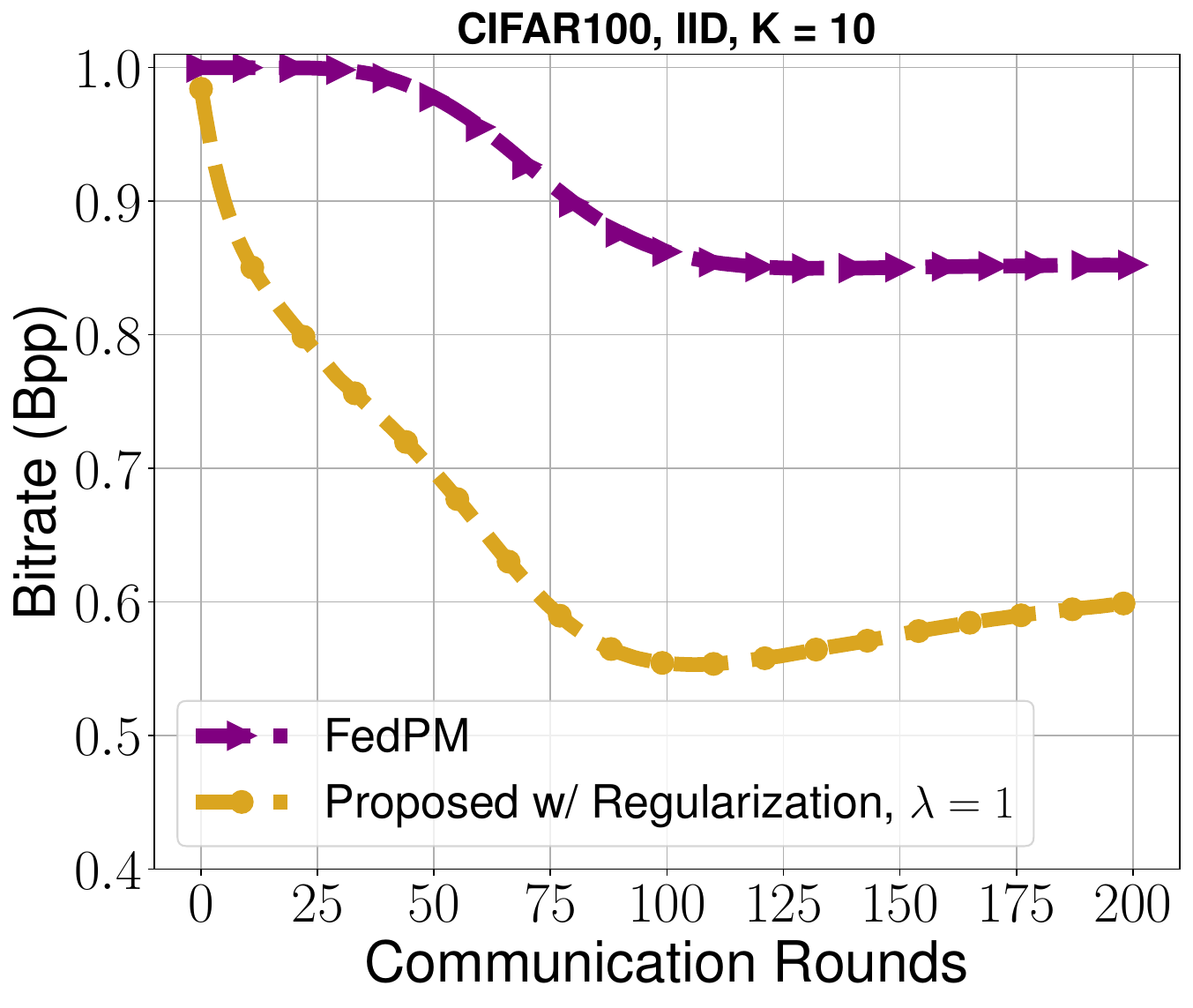}
    \captionsetup{labelformat=empty}
         \caption{(c) CIFAR100, IID}
         \label{fig:EMNIST_cov}
     \end{subfigure}
    \centering
     \caption{    \centering
From left to right: CIFAR10, MNIST, CIFAR100 experiments. First row: Validation Accuracy vs Rounds. Second row: The corresponding Average Bit-per-parameter (Bpp) required vs Rounds.}
     \label{acc}
\end{figure*}
Given the global mask $\boldsymbol{\theta}$ as defined in \eqref{global_mask}, which represents the average of the local masks transmitted in the uplink, the optimization of $H(\boldsymbol{m})$, the entropy of the global mask  $\boldsymbol{m}$, serves to minimize the average entropy across the set of local masks.
To optimize this new objective, a new definition of the local loss functions at each device $i$ over a mini-batch $\mathcal{B}\subseteq \mathcal{D}_i$ is introduced and is given by:
\begin{equation}
L_i(y_{{\boldsymbol{m}}_i}, \mathcal{B}) = \ell(y_{{\boldsymbol{m}}_i},  \mathcal{B}) + \frac{\lambda}{n} \sum_{j=1}^{n} \sigma(s_{i, j}).
\end{equation}
Where $n$ denotes the number of parameters in the over-parameterized network. The new loss includes a regularization term encompassing the local probability mask parameters $\theta_{i,j}=\sigma(s_{i,j})$ given in \eqref{probability_update} for each device $i$. The regularization term acts as proxy to the entropy of the local transmitted masks. The incorporated regularization term induces sparsity within the local masks by reducing the probability that redundant elements are retained (set to 1). As a result, $p_{0,k}$ is maximized, promoting a global mask with low entropy. Furthermore, this approach reduces the likelihood of sampling entirely new and distinct sub-networks in subsequent iterations, favoring the sampling of sub-networks sharing substantial features with early stages samples. Under regularization, upon transmitting the updated mask to the server, the resulting global probability mask defined in \eqref{global_mask} introduces redundant sub-networks once again due to the inherent stochasticity in the local sub-network sampling process on each device. However, this global mask also characterizes a more constrained search space for the devices in the subsequent rounds as the training progress, given the limited number of distinct sub-networks optimized by each device during successive iterations.
The training proceeds until a probability mask is found that can produce sub-networks with sparsity guided by the parameter $\lambda$, thereby ensuring both communication and memory efficiency, alongside achieving good generalization performance.

\section{Experiments}
To assess the effectiveness of our proposed approach, we carry out a series of experiments involving image classification tasks. These experiments are conducted under both homogeneous and heterogeneous conditions, as follows:
\begin{itemize}
    \item An Independent and Identically Distributed (IID) scenario, we evenly distribute the datasets CIFAR10, CIFAR100 \cite{ref10}, and MNIST \cite{ref11} across 10 devices.
    \item Non-IID setting where we distribute the MNIST and CIFAR10 datasets across 30 devices while introducing heterogeneity by randomly assigning each device a subset of $c=\{2,4\}$ classes from the available 10 classes.

\end{itemize}
For these experiments, we present the average testing accuracy over the devices target distribution (top row) and the average bits per parameter  required (lower row) as a function of the number of rounds. The reported average bits per parameter required represents the average estimated entropy of the binary source producing the binary masks transmitted in the UL by the devices during each round. The entropy can be written as:
\begin{equation}
\hat{H} = -\frac{1}{K}\sum^K_k\hat{p}_{k,0} \log (\hat{p}_{k,0})+ \hat{p}_{k,1} \log (\hat{p}_{k,1}) 
\end{equation}
where $\hat{p}_{k,0}$ and $\hat{p}_{k,1} = 1 - \hat{p}_{k,0}$ denote the  normalized frequencies of 0's and 1's in the observed mask at each device $k$ during each round.  Accordingly, $\hat{p}_{k,0}$ and $\hat{p}_{k,1}$ serve as an estimate of $p_{k,0}$ and $p_{k,1}$ respectively. The number of local epochs is set to three with $|\mathcal{B}|$ = 128. We utilize three feed-forward convolutional networks (4Conv, 6Conv and 10Conv as in \cite{ref9}) to train over MNIST, CIFAR10 and CIFAR100 respectively. We set $\ell$ to be the cross entropy loss. The initial global probability mask is sampled from a uniform distribution $\mathcal{U}[0,1]$. As in \cite{ref4,ref5,ref8}, the model random weights are sampled from a uniform distribution over $\{-\varsigma,\varsigma\}$, where $\varsigma$ denotes the standard deviation of the Kaiming Normal distribution.

\begin{figure}[ht!]
\centering
    \begin{subfigure}[t]{0.24\textwidth}
         \centering
          \includegraphics[width=\textwidth]{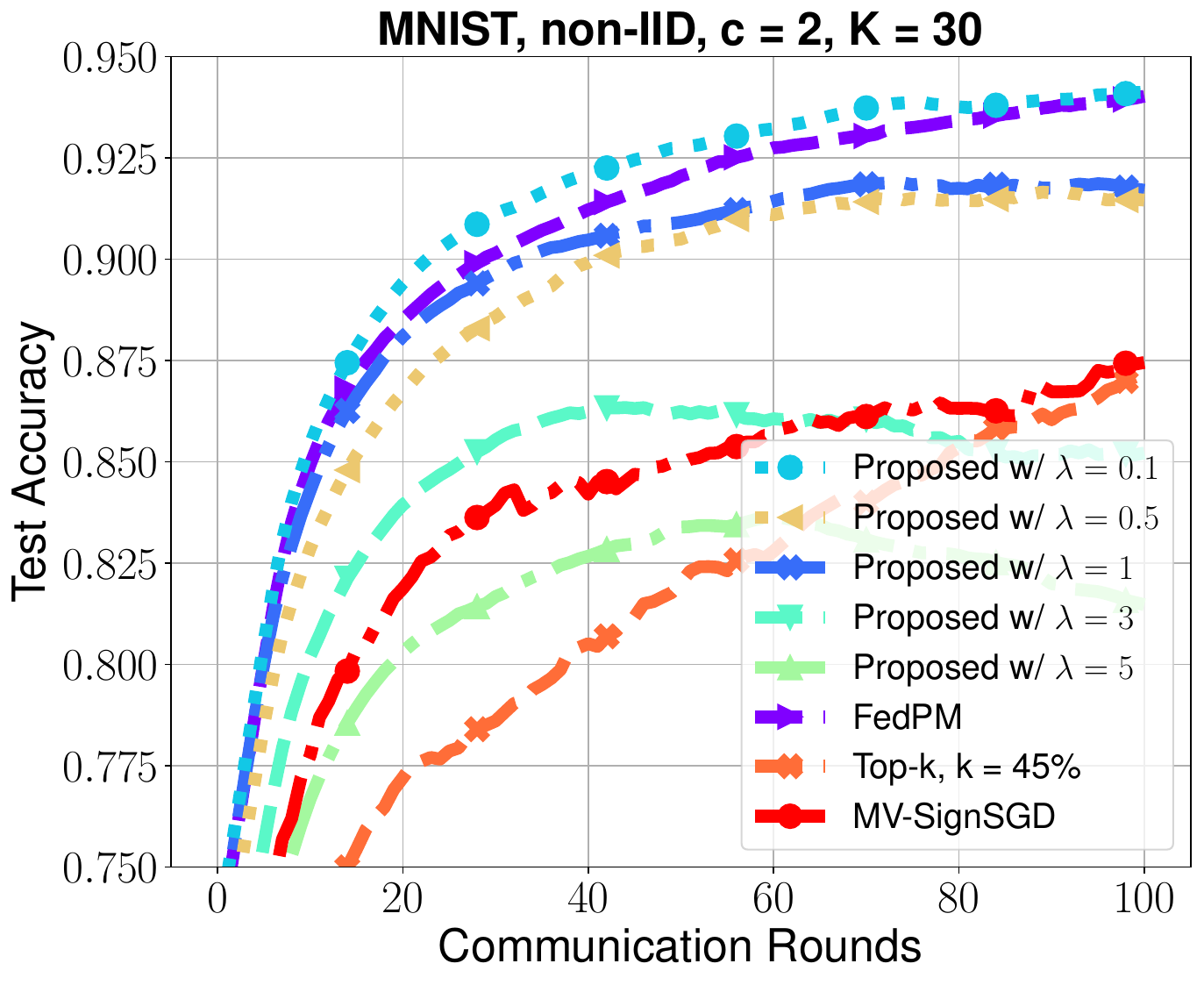}
         \centering
         \captionsetup{labelformat=empty}
         \caption{}
         \label{fig:optimized}
     \end{subfigure}
      \begin{subfigure}[t]{0.24\textwidth}
         \centering
          \includegraphics[width=\textwidth]{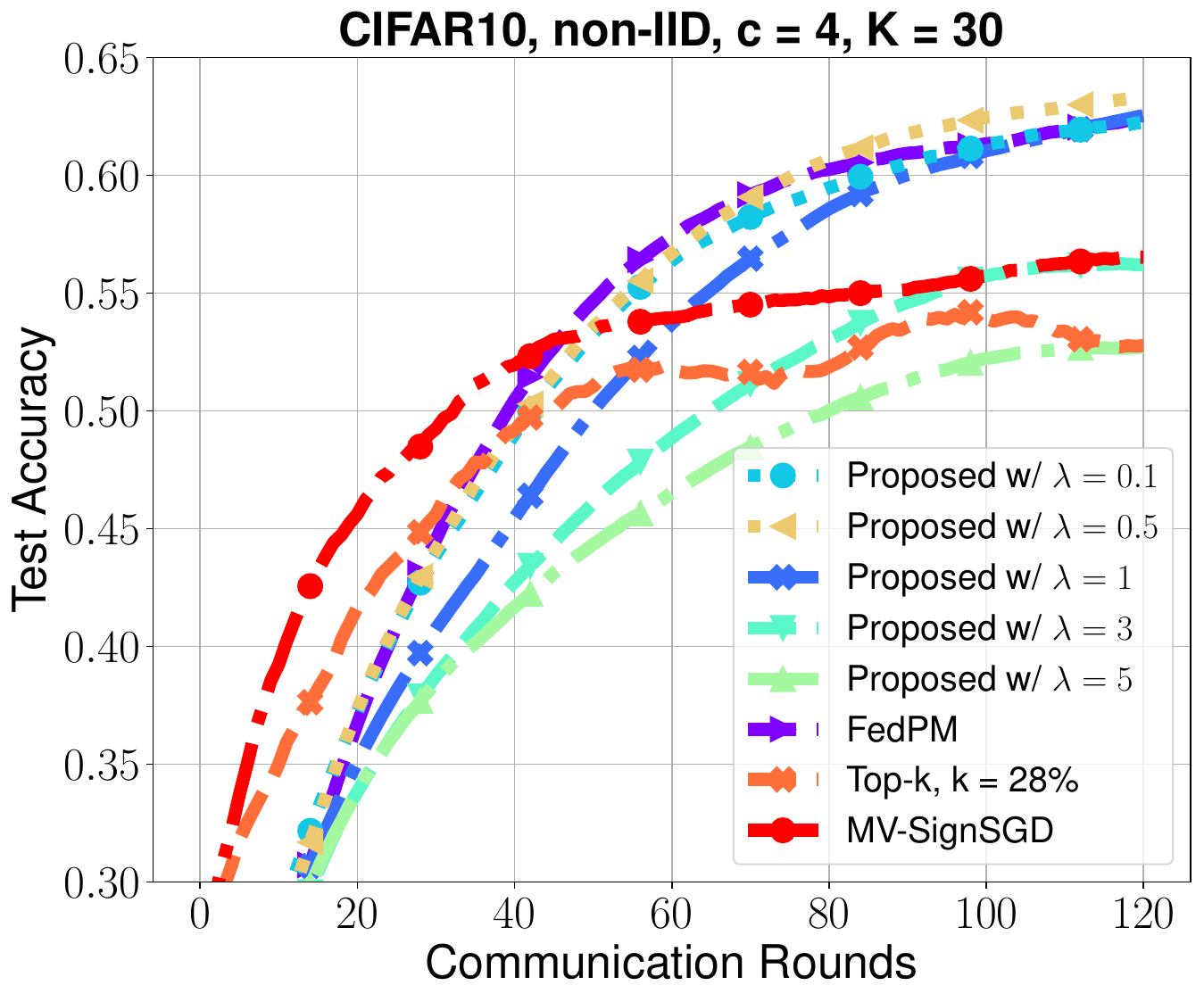}
         \captionsetup{labelformat=empty}
         \caption{}
         \label{fig:rect}
     \end{subfigure}

         \begin{subfigure}[t]{0.24\textwidth}
         \centering
            \includegraphics[width=\textwidth]{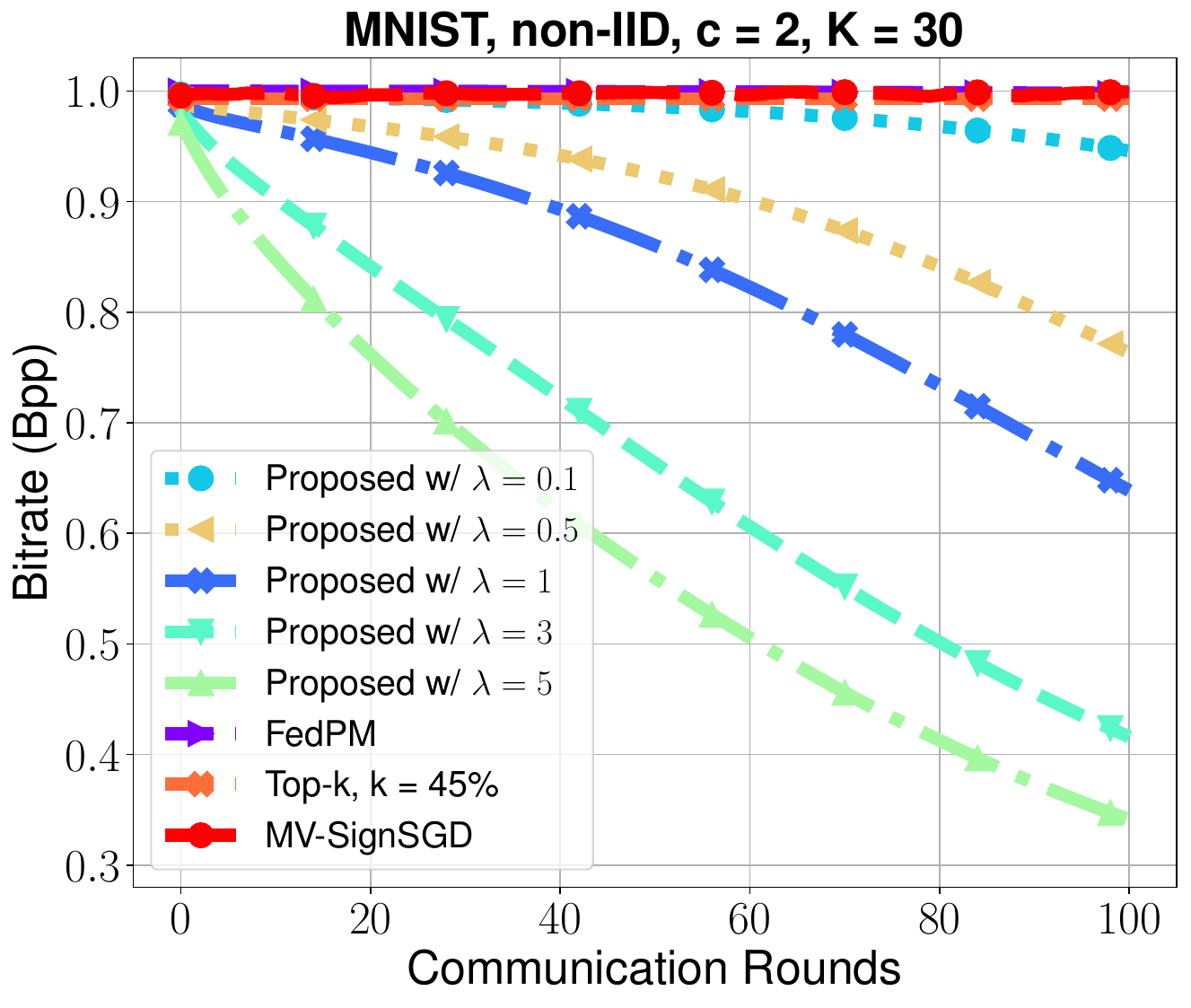}
         \centering
        \captionsetup{labelformat=empty}
         \caption{(a) MNIST, non-IID, $c=2$}
         
         \label{fig:optimized1}
     \end{subfigure}
      \begin{subfigure}[t]{0.24\textwidth}
         \centering
          \includegraphics[width=\textwidth]{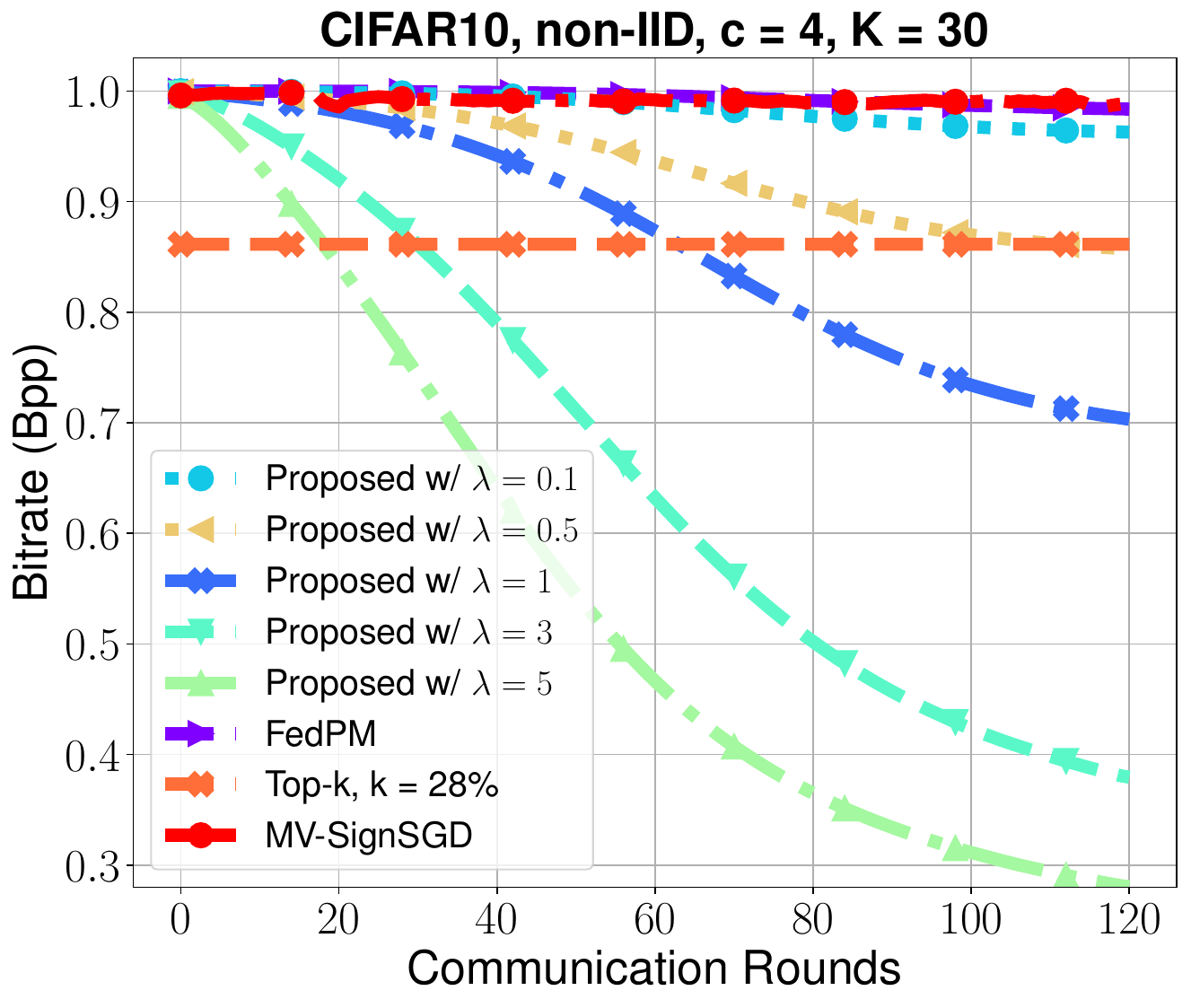}
         \captionsetup{labelformat=empty}
         \caption{(b) CIFAR10, non-IID, $c=4$}
         \label{fig:rect1}
     \end{subfigure}
         \caption{\centering Trade-off between validation accuracy and average Bpp for different regularization $\lambda$ in non-IID CIFAR10 and MNIST datasets settings. }
     
\end{figure}
Figure 1 illustrates the validation accuracy of FedPM with our proposed regularization term ($\lambda=1$) compared to the original algorithm under IID settings. The validation accuracy of both techniques is similar across all simulations. However, FedPM combined with our proposed regularization term achieves significant improvement in communication efficiency compared to the original algorithm. Specifically, on CIFAR10 experiments, an average efficiency gain of 0.31 bits per parameter (Bpp) is achieved using our proposed modification. On MNIST experiment, we achieve 0.8 Bpp greater efficiency, while on CIFAR100 experiment, we gain 0.25 Bpp higher efficiency relative to original algorithm. Therefore, our proposed recipe provides notable gains in communication efficiency while maintaining the generalization performance of FedPM in the IID settings configuration.

We now examine Fig. 2, which shows the performance of the proposed algorithm against FedPM and two other  baseline algorithms in the more challenging non-IID MNIST and CIFAR10 dataset settings. In particular, as a baseline, we add Top-k and Majority Vote SignSGD \cite{MV-sgd} algorithms. Top-k \cite{ref4} sets the top k\% of parameters in terms of their score, to 1, while pruning the rest. Majority Vote SignSGD is a variant of SignSGD for parallel training. The regularization term value of our proposed algorithm is varied to highlight the potential trade-off between accuracy and communication efficiency. In the MNIST experiment, for $\lambda=0.1$, the communication efficiency gain trend persists, where we observe marginal improvement of 0.05 bits per parameter (Bpp) compared to FedPM when label heterogeneity is present with $c=2$ while also enjoying a slight advantage in terms of convergence rate. As we increase the value of $\lambda$, the accuracy degrades, while witnessing higher communication saving gains. For instance, for $\lambda = 1$, average Bpp reduction of 0.35 compared to FedPM is observed while suffering around 2\% test accuracy loss. Top-k and MV-SignSGD are shown to suffer from poor generalization performance and higher communication costs. In the CIFAR10 experiment, in Fig. \ref{fig:rect}, we observe savings of 0.12 Bpp while enjoying a similar generalization performance as FedPM when $\lambda = 0.5$. Top-k algorithm converges faster initially, however suffers from poor performance in later stages, despite being assigned the same sparsity level of the sub-network obtained by the proposed algorithm for $\lambda = 0.5$. MV-SignSGD follows the same trend in terms of faster early convergence rate, and poor performance in later ones. It is noteworthy that while MV-SignSGD transmits binary masks during training, the final model necessitates storage in floating point representation. This goes in contrast to the strong LTH approach, which only requires a SEED and a binary mask to represent the attained sparse sub-network.

\section{Conclusion}
In this work, we demonstrate that state-of-the-art Federated learning training methods of random over-parameterized networks, which rely on consistent objectives, fail to uncover sparse sub-networks within the over-parameterized random models. To address this limitation, we propose and validate the incorporation of a regularization term within the local loss functions to discover highly sparse sub-networks. The sparse models obtained through our approach lead to significant improvements in communication and memory efficiency during federated training on resource-constrained edge devices, with minimal performance degradation. Through extensive experiments, we show that our method outperforms existing state-of-the-art techniques by a large margin in terms of the sparsity and efficiency gains achieved.



\bibliographystyle{IEEEtran}
\bibliography{biblio}

\vfill

\end{document}